\documentclass{article}
\usepackage{spconf,amsmath,graphicx}

\usepackage{enumitem}
\setlist{nosep, leftmargin=14pt}
\usepackage{url}

\usepackage{graphicx}
\usepackage{subcaption} % Recommended for modern LaTeX
\usepackage{adjustbox}
\usepackage{multirow} 
\usepackage{amsmath}

\usepackage{booktabs} % To thicken table lines
\usepackage{xcolor, colortbl}
\usepackage{capt-of}% or \usepackage{caption}
\usepackage{amssymb}
\usepackage{amsfonts}
\definecolor{cgreen}{rgb}{0,0.7,0.8}
\definecolor{cred}{rgb}{0.968,0.545,0.321}

\newcommand{\gtr}[1]{{\scriptsize\color{cgreen}$\blacktriangle$ #1}}
\usepackage{mwe} % to get dummy images

\title{LocBAM: Advancing 3D Patch-Based Image Segmentation by Integrating Location Context}
\name{Donnate Hooft$^{\star}$\thanks{$^{\star}$Shared First Author}$^{1}$, Stefan M. Fischer$^{\star}$$^{1,2,3,4}$, Cosmin Bercea$^{1,4}$, Jan C. Peeken$^{2}$, Julia A. Schnabel$^{1,3,4,5}$}
\address{$^{1}$ School of Computation, Information and Technology, Technical University Munich, Germany \\
$^{2}$ Department of Radiation Oncology, School of Medicine and Klinikum Rechts der Isar \\
$^{3}$ Institute of Machine Learning in Biomedical Imaging, Helmholtz Munich \\
$^{4}$ Munich Center of Machine Learning (MCML) \\
$^{5}$ School of Biomedical Engineering and Imaging Sciences, King's College London, UK}

\begin{document}
\ninept
\maketitle

Patch-based methods are widely used in 3D medical image segmentation to address memory constraints in processing high-resolution volumetric data. However, these approaches often neglect the patch’s location within the global volume, which can limit segmentation performance when anatomical context is important. In this paper, we investigate the role of location context in patch-based 3D segmentation and propose a novel attention mechanism, LocBAM, that explicitly processes spatial information. Experiments on BTCV, AMOS22, and KiTS23 demonstrate that incorporating location context stabilizes training and improves segmentation performance, particularly under low patch-to-volume coverage where global context is missing. Furthermore, LocBAM consistently outperforms classical coordinate encoding via CoordConv. Code is publicly available at~\url{https://github.com/compai-lab/2026-ISBI-hooft}.

\section{Introduction}
Deep learning has become the standard approach for 3D medical image segmentation. However, volumetric CT and MRI scans pose substantial memory challenges, especially at high resolutions. Patch-based strategies mitigate this by subdividing large volumes into smaller patches that fit GPU memory, balancing computational costs with anatomical context \cite{isensee2021nnu}.  

Even though the spatial location of each patch within the full scan is known, it is mostly ignored during training. In contrast, vision transformers leverage positional embeddings to encode spatial relationships across tokens \cite{alexey2020image}. Extending this principle to patch-based segmentation suggests that explicitly incorporating patch location can provide valuable anatomical priors. Previous work has shown that location context can significantly improve segmentation \cite{wang2019automated,ulrich2024mitigating,wang2020segmentation,rachmadi2018segmentation,breznik2023introducing,das2024co,el2021coordconv}. Methods differ primarily in: (1) how location context is represented and (2) how it is integrated into models. Common representations include voxel coordinates \cite{wang2019automated,el2021coordconv}, scanner coordinates \cite{das2024co}, anatomical landmarks \cite{breznik2023introducing}, or normalized body position scores from regression models \cite{yan2018unsupervised, schuhegger2021body}. Some priors might only be effective in well-aligned fields-of-view (FOV), e.g., atlas-registered brain MRIs \cite{wang2020segmentation} or cropped lungs \cite{wang2019automated}, but performance degrades under high FOV variability, such as in MSD Spleen segmentation \cite{el2021coordconv}.  

% Integration strategies also vary. Most CNN-based methods inject location as extra input channels \cite{wang2019automated,rachmadi2018segmentation,breznik2023introducing}, but CoordConv layers also allow injection at intermediate layers \cite{el2021coordconv}. Transformers, on the other hand, can incorporate spatial priors as positional embeddings \cite{das2024co}. More advanced strategies include anatomical priors via loss functions \cite{ulrich2024mitigating} or spatially-aware attention mechanisms, as in HANet for urban scene segmentation \cite{choi2020cars}.  

Amongst the different integration strategies are adding location as extra input channels in CNNs \cite{wang2019automated,rachmadi2018segmentation,breznik2023introducing}, injecting it at intermediate layers via CoordConv \cite{el2021coordconv}, encoding spatial priors as positional embeddings in Transformers \cite{das2024co}, and employing advanced priors through loss functions \cite{ulrich2024mitigating} or spatially-aware attention mechanisms such as HANet \cite{choi2020cars}.

Motivated by these findings, we hypothesize that patch-based segmentation can benefit from location context, especially with limited patch-to-volume coverage. Since anatomical structures and pathologies follow predictable spatial distributions, such priors can enhance robustness and discrimination. Our contributions are:
\begin{itemize}
\item We systematically evaluate location context in patch-based 3D medical image segmentation across organ and lesion tasks.
\item We propose LocBAM, a lightweight 3D attention mechanism for integrating location context into patch-based models.
\item We demonstrate that LocBAM outperforms CoordConv on BTCV, AMOS22, and KiTS23, remains robust to spatial distortions, and improves stability with larger patch-to-volume ratios.
\end{itemize}

\begin{figure*}[th!]
    \centering
    \includegraphics[width=0.7\textwidth]{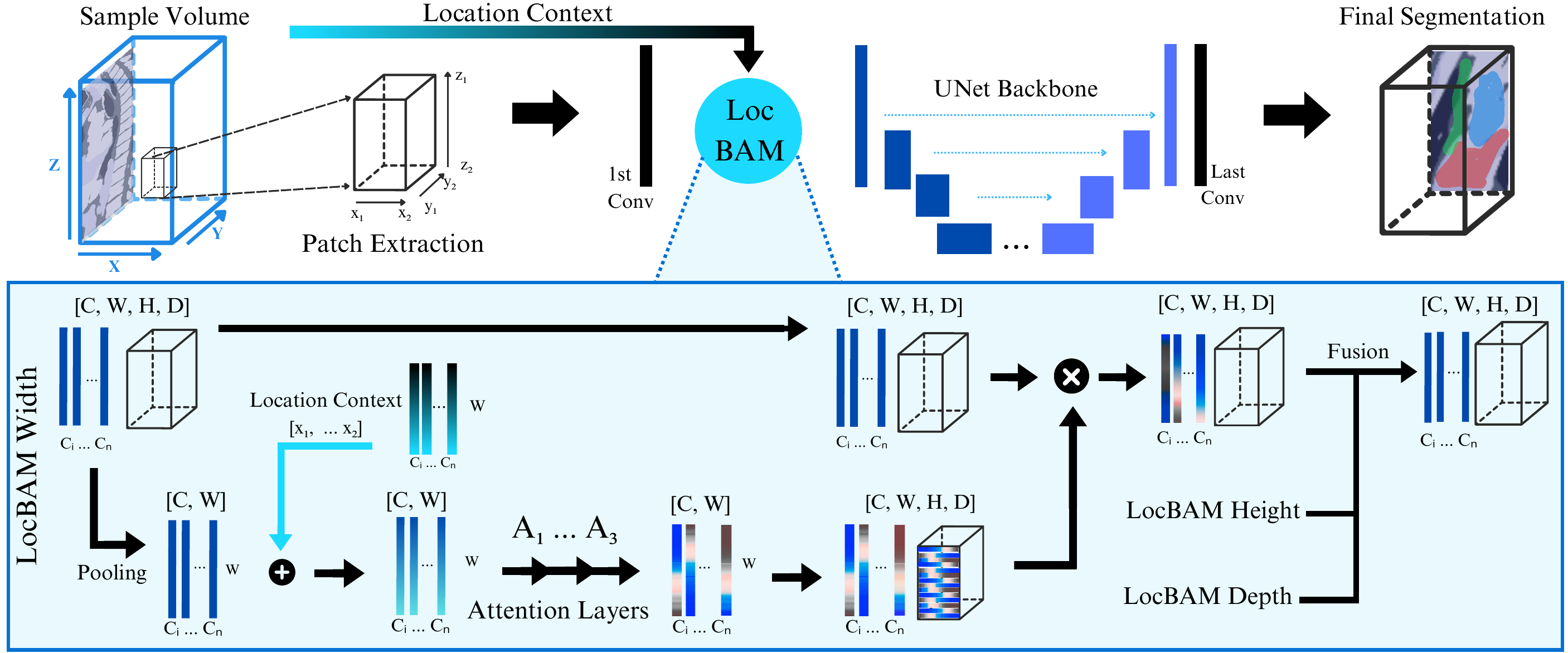}
    \caption{LocBAM integrates location context into a (patch-based) segmentation backbone. Overview of LoCBAM integrated into the second block of a 3D U-Net. LocBAM extends HANet’s \cite{choi2020cars} mechanism to 3D by applying independent 1D attention gates along width, height, and depth, then fusing their outputs via a $1\times1$ convolution.} 
    \label{fig:combined}
\end{figure*}
\section{Methods}
We introduce LocBAM, a memory-efficient attention mechanism for incorporating spatial context into 3D patch-based medical image segmentation. As baselines, we compare against location encoding via CoordConv \cite{liu2018intriguing}, which has been adopted in medical segmentation \cite{ulrich2024mitigating, wang2020segmentation, rachmadi2018segmentation, breznik2023introducing, el2021coordconv}, and against classical postprocessing strategies.  

To normalize anatomical position, we apply Body Part Regression (BPR) \cite{schuhegger2021body}, assigning axial CT slices a 0–100 score from pelvis to head.

As additional baselines, we evaluate two classical postprocessing techniques: \textbf{Largest Component Filtering (LCF)} \cite{isensee2021nnu}, which removes small spurious components, and \textbf{Atlas Masking} \cite{engelson2024lnq}, which eliminates predictions in implausible regions. For BTCV, where all scans share a common FOV, we built an atlas by resampling all labels to the median volume size, averaging them, and applying dilation (kernel optimized on validation Dice).  

\subsection{Location-Based Attention Mechanism (LocBAM)}
LocBAM extends the hierarchical attention concept from HANet \cite{choi2020cars} to 3D patch-based segmentation. Unlike CoordConv, which directly encodes coordinates as input channels, LocBAM learns to weight features based on spatial position.  

For memory efficiency, LocBAM applies three independent 1D convolutional attention gates (HANet's attention mechanism) along width, height, and depth. The resulting activations are concatenated and fused with a $1\times1$ convolution. Following HANet’s integration strategy, LocBAM is placed in the second convolutional block of a standard 3D U-Net (Fig.~\ref{fig:combined}). Hyperparameters are taken directly from HANet’s default configuration.

\begin{table}[b!]
\centering
\small
\caption{Benchmarking of location-context integration methods. Average Dice Score (\%) for BTCV, AMOS22, and KiTS23 at 1mm isotropic spacing (patch size $128^3$, batch size 2).}
\begin{tabular}{l c c c}
\hline
\textbf{Model} & \textbf{BTCV} & \textbf{AMOS22-CT} & \textbf{KiTS23} \\ 
\hline
Baseline & 90.27 & 87.28 & 80.85 \\
CoordConv & 86.42 & 87.43 & 80.64 \\
LocBAM (Ours) & \textbf{90.66} & \textbf{87.44} & \textbf{81.39} \\
\hline
\end{tabular}
\label{tab:datasets_dice_scores}
\end{table}

\section{Experiments and Results}

We evaluate the impact of incorporating location context on three publicly available 3D medical imaging datasets: BTCV-Abdomen \cite{landman2015miccai}, AMOS22-CT \cite{ji2022amos}, and KiTS23 \cite{heller2023kits21}. These datasets cover diverse segmentation challenges: multi-organ abdominal segmentation (BTCV, AMOS22-CT) and kidney tumor segmentation (KiTS23). All experiments were implemented in MONAI, following the nnU-Net training protocol \cite{isensee2021nnu}, but excluding data augmentation to isolate the effect of location context.  

\subsection{Benchmarking}

\textbf{Setup.}  
We benchmark LocBAM against CoordConv across all datasets. Since AMOS22 and KiTS23 exhibit varying fields-of-view (FOVs), we replaced raw axial coordinates with Body Part Regression (BPR) scores to provide a normalized anatomical axis. Values outside the pelvis–head range were linearly extrapolated. Nine KiTS23 scans with corrupted orientation or spacing were excluded. All datasets were resampled to 1mm isotropic resolution, using a patch size of $128^3$ and batch size 2.  
\\
\textbf{Results.}  LocBAM consistently improved segmentation performance across datasets (Table~\ref{tab:datasets_dice_scores}). In contrast, CoordConv yielded mixed results: slight improvement on AMOS22 (+0.17\%) but substantial degradation on BTCV (–4.26\%) and KiTS23 (–0.26\%). Overall, LocBAM outperformed both the baseline and CoordConv in all cases, with gains of up to +0.67\% Dice on KiTS23 and +0.43\% on BTCV.  

\begin{table}[b!]
\centering
\caption{ Average Dice Score (\%) on BTCV across patch-to-volume coverage settings and spacings. Comparison between Baseline and LocBAM (ours).}
\label{tab::performance}
\small
\begin{tabular}{l l l l}
\toprule
\textbf{Type} & \textbf{PtVC} & \textbf{Model} & \textbf{Dice $\uparrow$} \\ 
\hline
\multirow{4}{*}{Low-Res} & \multirow{2}{*}{0.26\%} & Baseline & 92.02 \\ 
& & LocBAM & 93.90~\gtr{2.04\%} \\ 
&  \multirow{2}{*}{16.67\%} & Baseline & 93.05 \\
& & LocBAM & 93.09~\gtr{0.04\%} \\\hline 
\multirow{4}{*}{Isotropic} & \multirow{2}{*}{0.06\%} & Baseline & 36.15 \\ 
& & LocBAM & 91.40~\gtr{152.84\%} \\ 
&  \multirow{2}{*}{3.70\%} & Baseline & 90.27 \\
& & LocBAM & 90.66~\gtr{0.43\%} \\\hline 
\multirow{4}{*}{Median} & \multirow{2}{*}{0.10\%} & Baseline & 54.72 \\ 
& & LocBAM & 91.80~\gtr{67.76\%} \\ 
&  \multirow{2}{*}{5.27\%} & Baseline & 85.52 \\
& & LocBAM & 88.87~\gtr{3.92\%} \\ 
\bottomrule
\end{tabular}
\end{table}

\subsection{Patch-to-Volume Coverage}

\textbf{Setup.}  
We investigated how patch-to-volume coverage (PtVC) affects the benefit of location context on BTCV. Three commonly used spacings in 3D patch-based segmentation were examined: (i) nnU-Net’s median spacing for BTCV (3.0, 0.75, 0.75) \cite{isensee2021nnu}, (ii) isotropic 1 mm spacing \cite{hatamizadeh2022unetr}, and (iii) a low-resolution spacing (2.0, 1.5, 1.5) \cite{he2023swinunetr}. For each spacing, we trained with both large patches ($128^3$, batch size 2) and small patches ($32^3$, batch size 128) to modulate global context. For median spacing, we adopted nnU-Net’s standard large patch setup ($48 \times 192 \times 192$, batch size 2).

\begin{figure}[t!]
    \centering
    \includegraphics[width=\linewidth]{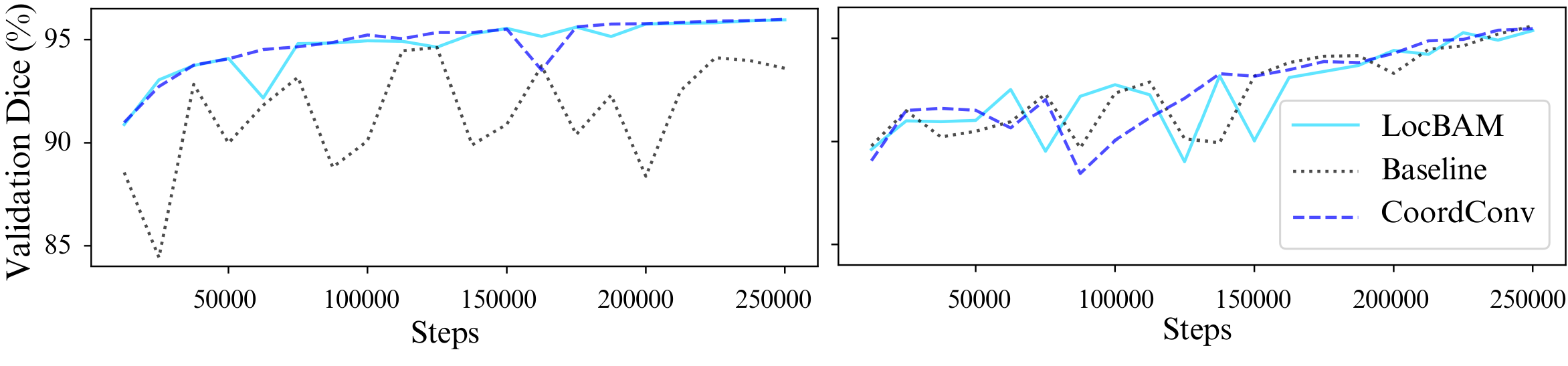}
    \caption{Training curves for BTCV at low-resolution with small patches (Left) and large patches (Right). LocBAM stabilizes performance in low coverage scenarios.}
    \label{fig::stability}
\end{figure}

\textbf{Results.}  
Across all settings, LocBAM improved segmentation performance compared to the baseline (Table~\ref{tab::performance}). The relative gain was largest in scenarios with limited PtVC. In high-resolution settings (isotropic, median spacing), small patch models without location context often failed to converge, while LocBAM maintained stable training. The best overall performance was achieved at low-resolution with small patches, where LocBAM provided substantial stability gains (Fig.~\ref{fig::stability}). Unlike the baseline, which showed oscillating validation Dice scores, LocBAM and CoordConv stabilized convergence, with LocBAM achieving consistently higher scores. Since the low-resolution small-patch setting was the only scenario that converged without location context (Table \ref{tab::performance}), we focused further analysis on this case when comparing patch sizes, alternative methods, and location shifts.

% The low-resolution small-patch setting was selected for further analysis, given that it was the only scenario to achieve convergence in the absence of location context.

\subsection{Classical Postprocessing vs. Location Context}

\textbf{Setup.}  
We compared classical postprocessing methods (largest component filtering, atlas masking) with location-aware models (CoordConv, LocBAM) in the low-resolution, small-patch BTCV setting.
% This was the only scenario in which model training converged with small patch sizes, as shown in Table \ref{tab::performance}.

\begin{figure}[b!]
    \centering
    \includegraphics[width=0.45\textwidth]{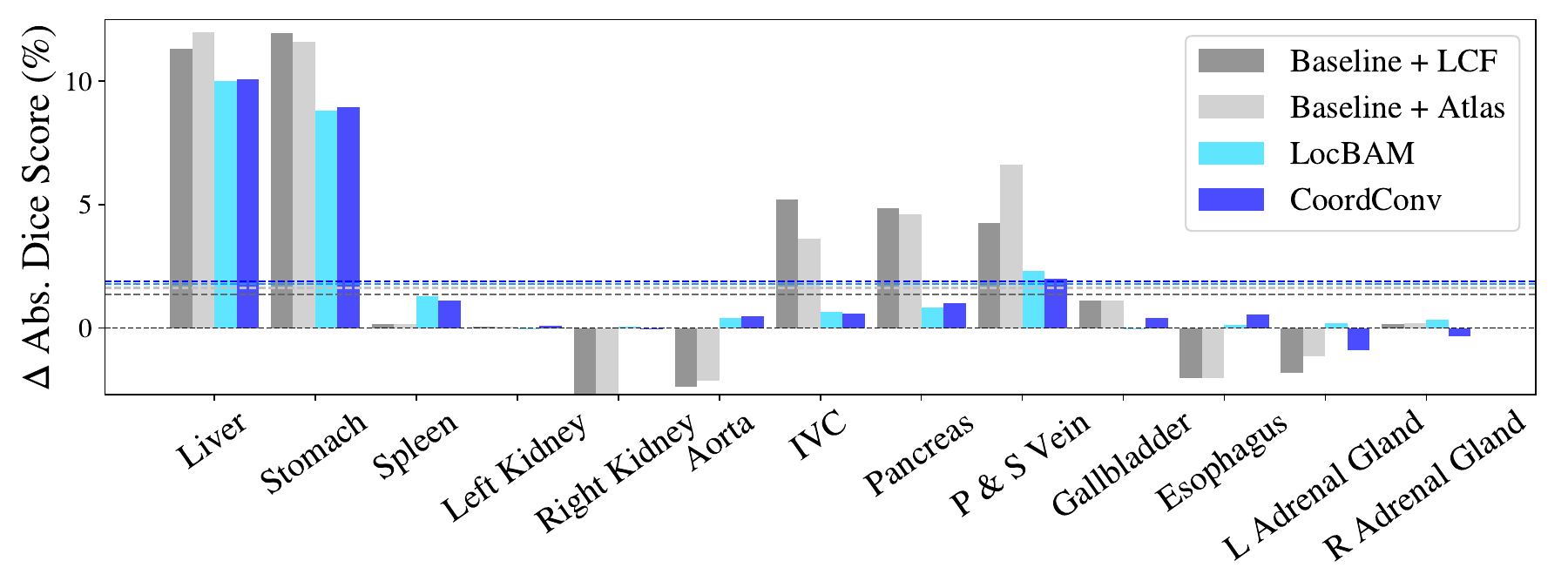}
    \caption{Class-wise Dice score improvements over baseline on BTCV. Classes are ordered by size. Dotted lines indicate average gains per method. LocBAM improves performance consistently, especially for large organs.}
    \label{fig:dice_class_wise_ordered}
\end{figure}

\textbf{Results.}  
The baseline achieved 92.02\% mean Dice. Postprocessing improved performance to 93.37\% (largest component filtering) and 93.65\% (atlas masking). Location-aware models outperformed both approaches, with CoordConv reaching 93.81\% and LocBAM 93.90\%. Class-wise analysis (Fig.~\ref{fig:dice_class_wise_ordered}) revealed that while postprocessing and CoordConv occasionally reduced performance for smaller classes, LocBAM consistently improved or preserved performance across all organs.

\subsection{Robustness to Noisy Location Information}

\textbf{Setup.}  
To evaluate robustness, we systematically perturbed the location signals provided to location-aware models by introducing axial shifts of varying magnitudes (0\%, 1\%, 5\%, 10\%, 25\%, 50\%, 100\%), where 100\% corresponds to a completely out-of-frame position. Experiments were conducted on BTCV in the low-resolution setting for both small and large patch sizes.  

\begin{figure}[!t]
    \centering
    \begin{subfigure}[t]{0.23\textwidth}
        \centering
        \includegraphics[width=\textwidth]{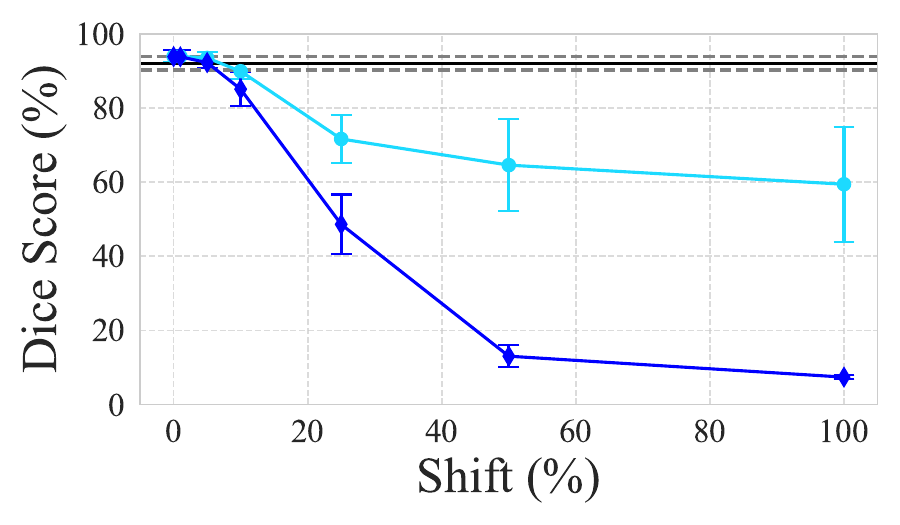}
        \caption{Small patches}
    \end{subfigure}
    \hfill
    \begin{subfigure}[t]{0.23\textwidth}
        \centering
        \includegraphics[width=\textwidth]{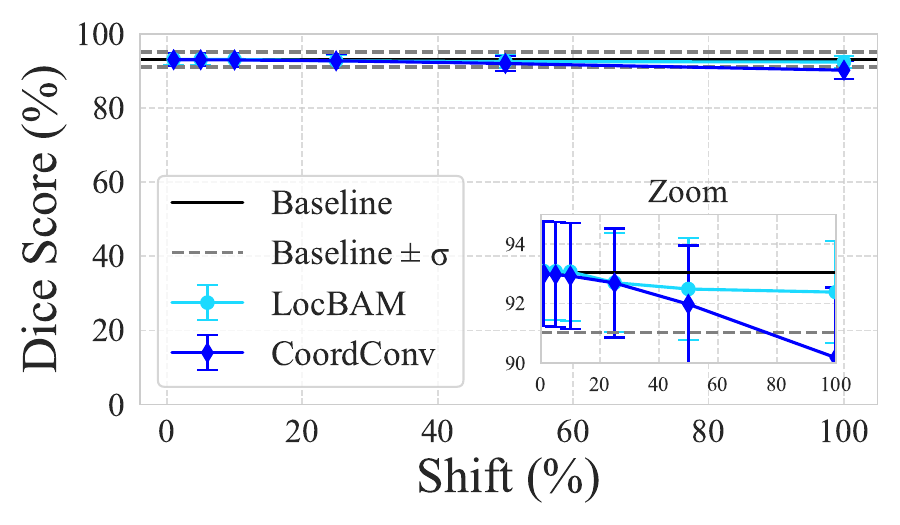}
        \caption{Large patches}
    \end{subfigure}
    \caption{Sensitivity of CoordConv and LocBAM to axial location shifts on BTCV. For small patches, CoordConv degrades sharply while LocBAM remains more stable. Large-patch models are minimally affected.}
    \label{fig:ablate}
\end{figure}

\textbf{Results.}  
In the small-patch setting, both CoordConv and LocBAM exhibited performance degradation as shift magnitude increased (Fig.~\ref{fig:ablate}). However, LocBAM was substantially more robust, maintaining Dice scores above 59.4\% even at 100\% shifts, whereas CoordConv dropped to 7.39\%. For large patches, performance remained largely unaffected by noisy location information, with LocBAM showing the smallest variation across shifts. Minor perturbations (1–5\%) had negligible impact in all cases.

\section{Discussion and Conclusion}

This study demonstrates that incorporating location context markedly improves patch-based medical image segmentation across diverse tasks. We introduced LocBAM, the first 3-dimensional attention-based approach to spatial encoding in medical image segmentation. Unlike CoordConv, which may degrade performance in datasets with large PtVC or under noisy location information, LocBAM integrates location context in a memory-efficient and robust manner. Our experiments show that location context integration (LocBAM, CoordConv) not only outperforms the baseline but also stabilizes training in challenging small-PtVC regimes.

Across BTCV, AMOS22, and KiTS23, the proposed LocBAM mechanism consistently enhanced segmentation performance. The largest gains occurred under extremely limited patch-to-volume coverage (PtVC), with up to a 152\% Dice improvement at 0.06\% PtVC, indicating that location context becomes increasingly critical as PtVC decreases. Even under moderate PtVC conditions (e.g., 6\% in the nnU-Net BTCV setup), LocBAM achieved significant improvements of up to 4\%.

Additionally, LocBAM avoids negative class-wise effects observed when integrating postprocessing approaches such as Largest Component Filtering or Atlas masking. Larger classes benefited the most, but LocBAM was the only method that improved or maintained performance across all classes.

Taken together, these results highlight the broader value of location context for patch-based segmentation, especially in high-resolution data, variable FOVs, or constrained PtVCs. Integrating location context via LocBAM enabled small PtVC models to achieve performance comparable to large PtVC configurations. This indicates that global anatomical location is the key form of contextual supervision missing when using small PtVC. By reintroducing this information, LocBAM allows greater flexibility in selecting batch size and patch size than the baseline, reducing the need to rely solely on large patches for global context.
% --> cocnise w gtp

\textbf{Outlook.}  
This work establishes LocBAM as a proof of concept for integrating spatial priors through attention. Future work could extend regressed location encoding beyond the axial direction by developing analogous regressors for coronal and sagittal axes, thereby refining anatomical positioning. Data augmentation of location signals could further improve robustness by modeling inter-patient variability and positional uncertainty. As imaging resolutions increase, leading to smaller feasible PtVCs under memory constraints, the need for reliable spatial encoding will only grow. We therefore anticipate attention-based spatial integration, as exemplified by LocBAM, to play an increasingly important role in the next generation of 3D patch-based medical image segmentation models.


\section*{Compliance with Ethical Standards}
The study retrospectively used open-access human data, with no ethical approval required per the dataset license.

\section*{Acknowledgments}
SMF, JCP, and JAS acknowledge funding by the Deutsche Forschungsgemeinschaft (DFG, German Research Foundation) – 515279324 / SPP 2177. SMF also acknowledges funding via the EVUK program ("Next-generation Al for Integrated Diagnostics”) of the Free State of Bavaria.

\section*{Conflicts of Interest}

\emph{The authors have no relevant conflict of interest to disclose.}

\bibliographystyle{IEEEbib}
\bibliography{refs}

\end{document}